\documentclass[runningheads]{llncs}

\usepackage{graphicx}
\usepackage{float}
\usepackage{booktabs} 
\usepackage{multirow} 
\usepackage{colortbl}   
\usepackage{xcolor}     
\usepackage{graphicx}   
\usepackage{color}
\usepackage{amsmath}
\usepackage{amssymb}
\usepackage[T1]{fontenc}
\usepackage{graphicx,verbatim}
\usepackage{color}
\usepackage{bm}
\usepackage{subcaption}
\begin{document}
\title{Fake It Right: Injecting Anatomical Logic into Synthetic Supervised Pre-training for Medical Segmentation}

\author{Jiaqi Tang\inst{1}\thanks{These authors contributed equally to this work.} \and
Mengyan Zheng\inst{2}\protect\footnotemark[1] \and
Shu Zhang\inst{3} \and
Fandong Zhang\inst{3} \and
Qingchao Chen\inst{1}\thanks{Corresponding author.}}

\authorrunning{J. Tang et al.}

\institute{Peking University \\
\email{jiaqi\_tang@hsc.pku.edu.cn, qingchao.chen@pku.edu.cn} \and
Beihang University \\
\email{zmy050516@buaa.edu.cn} \and
Deepwise Ltd. \\
\email{\{zhangshu, zhangfandong\}@deepwise.com}}

\maketitle              

\begin{abstract}
Vision Transformers (ViTs) excel in 3D medical segmentation but require massive annotated datasets. While Self-Supervised Learning (SSL) mitigates this using unlabeled data, it still faces strict privacy and logistical barriers. Formula-Driven Supervised Learning (FDSL) offers a privacy-preserving alternative by pre-training on synthetic mathematical primitives. However, a critical semantic gap limits its efficacy: generic shapes lack the morphological fidelity, fixed spatial layouts, and inter-organ relationships of real anatomy, preventing models from learning essential global structural priors. 
To bridge this gap, we propose an \textbf{Anatomy-Informed Synthetic Supervised Pre-training} framework unifying FDSL's infinite scalability with anatomical realism. We replace basic primitives with a lightweight shape bank with de-identified, label-only segmentation masks from 5 subjects. Furthermore, we introduce a structure-aware sequential placement strategy to govern the patch synthesis process. Instead of random placement, we enforce physiological plausibility using spatial anchors for correct localization and a topological graph to manage inter-organ interactions (e.g., preventing impossible overlaps). 
Extensive experiments on BTCV and MSD datasets demonstrate that our method significantly outperforms state-of-the-art FDSL baselines and SSL methods by 1.74\% and up to 1.66\%, while exhibiting a robust scaling effect where performance improves with increased synthetic data volume. This provides a data-efficient, privacy-compliant solution for medical segmentation. The code will be made publicly available upon acceptance.

\end{abstract}

\section{Introduction}

Accurate 3D medical image segmentation is fundamental to clinical diagnosis and treatment planning. While Vision Transformers (ViTs), such as UNETR~\cite{hatamizadeh2022unetr} and SwinUNETR~\cite{hatamizadeh2022swinunetr}, have demonstrated impressive performance by capturing long-range dependencies, they lack the inductive biases inherent to CNNs~\cite{dosovitskiy2020image}. Consequently, these architectures are notoriously data-hungry and prone to overfitting when trained on limited annotations. This creates a critical bottleneck: collecting large-scale, voxel-wise annotated medical datasets is prohibitively expensive, and sharing even unlabeled data is severely restricted by privacy regulations and institutional silos.

To mitigate the dependency on annotated data, \textit{Self-Supervised Learning (SSL)} has emerged as a dominant paradigm. Methods utilizing Masked Image Modeling~\cite{Tang2022SwinUNETR,he2021mae,xie2022simmim} or contrastive learning~\cite{wu2024voco,zhou2019models} leverage vast archives of unlabeled scans to learn robust representations. However, SSL faces two intrinsic limitations. First, it does not circumvent the logistical and legal barriers of data curation, which makes models still require access to large-scale in-domain medical archives. Second, standard SSL objectives, such as intensity reconstruction, primarily focus on local features, often failing to provide explicit supervision.
Alternatively, Formula-Driven Supervised Learning (FDSL)~\cite{kataoka2021formula,shinoda2023segrcdb,takashima2023visual,tadokoro2023pre,nakamura2023pre,kataoka2022replacing}, offers a privacy-compliant and infinitely scalable solution. By procedurally generating image-label pairs from mathematical rules, FDSL enables the initialization of segmentation models without exposing them to any real patient data. In medical domain, prior works such as PrimGeoSeg~\cite{tadokoro2023pre} constructs 3D volumes by randomly arranging primitive geometric objects (e.g. polygons, cylinders). 
Nevertheless, a critical \emph{semantic gap} remains between these mathematical primitives and human anatomy. Existing FDSL methods rely on stochastic placement and simplified geometry, failing to capture the \textbf{morphological fidelity and fixed topological constraints} inherent in medical scans. For instance, \textit{in a FDSL-generated volume, a cylinder might be randomly positioned above a ``lung-like'' space, resulting in a spatial configuration that is topologically impossible in human physiology.} Consequently, models pre-trained on such chaotic synthetic data may master low-level edge detection but fail to acquire the essential \textbf{anatomical priors}, hindering their ability to distinguish soft tissues with low contrast where spatial context is as important.

To address these limitations, we introduce a hybrid paradigm that merges the infinite scalability of FDSL with the biological validity of real data. We propose an \textbf{Anatomy-Informed Synthetic Supervised Pre-training framework}. Unlike SSL approaches that rely on texture reconstruction or generic FDSL methods limited to geometric primitives, our paradigm leverages dense anatomical supervision derived from a generative composition process.
To be specific, we introduce a \textbf{structure-aware synthesis strategy} by constructing a lightweight shape bank from a minimal set of subjects ($K=5$) containing only geometric masks, discarding all patient-specific texture information.
Instead of random placement, we inject explicit anatomical priors, including class-wise spatial anchors and a topological relation graph to govern the generation. This compels the Transformer to learn valid global anatomical layouts and inter-organ dependencies from scratch, providing strong, pixel-level supervision that is completely privacy-preserving. We systematically evaluate our framework on BTCV~\cite{landman2015miccai} and MSD~\cite{antonelli2022medical} datasets on UNETR~\cite{hatamizadeh2022unetr} and SwinUNETR~\cite{hatamizadeh2022swinunetr} backbones, outperforming state-of-the-art FDSL baselines and SSL methods by over 1.45\% and up to 1.67\%.  

In summary, we present a novel pre-training paradigm that provides dense, task-aligned supervision without real data. By capturing global structural dependencies unavailable to standard FDSL and demonstrating a continuous scaling effect, our method establishes a robust, privacy-compliant path for data-efficient medical AI.

\begin{figure*}[t]
    \includegraphics[width=0.99\linewidth, height=0.56\linewidth]{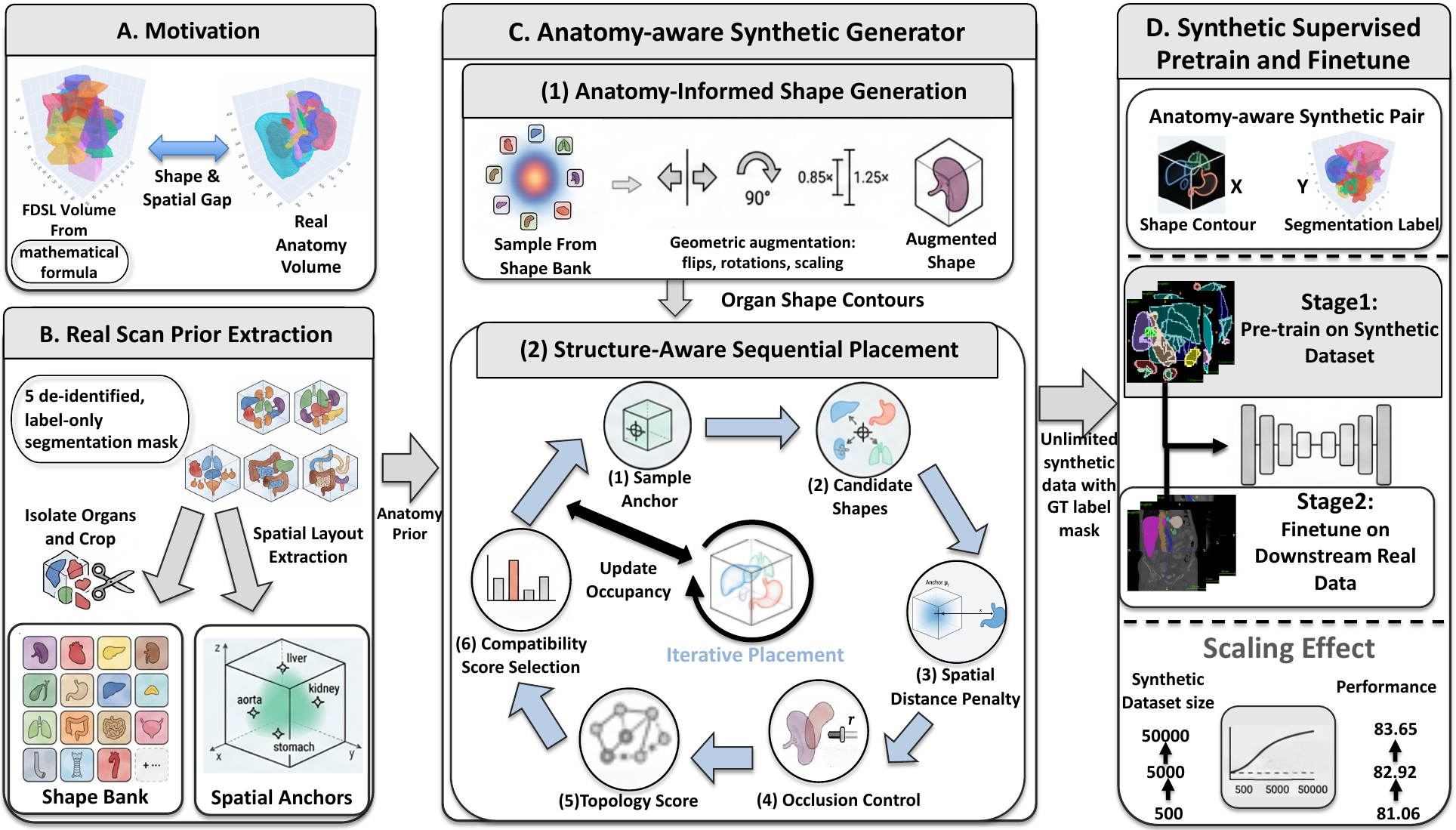}

    \caption{Overall framework.  By extracting morphological and spatial priors from minimal real masks, our generator iteratively synthesizes unlimited, biologically plausible data to bridge the semantic gap of FDSL for robust model pre-training and downstream fine-tuning.}
    \label{fig:framework}
    
\end{figure*}

\section{Method}

Figure \ref{fig:framework} illustrates our Anatomy-Informed Synthetic Supervised Pre-training framework, which generates biologically plausible 3D image-label pairs without relying on real patient textures. The pipeline consists of two core components: an anatomy-informed shape bank providing realistic morphological primitives (Sec.~\ref{sec:primitives}), and a structure-aware sequential placement mechanism governed by spatial and topological constraints (Sec.~\ref{sec:ranking}). By mathematically formulating this generation as a constrained spatial point process (Sec.~\ref{sec:formulation}), our method synthesizes diverse, densely annotated volumes to provide robust structural priors for segmentation model pre-training.

\subsection{Theoretical Formulation}
\label{sec:formulation}

The fundamental objective of synthetic pre-training is to learn a segmentation function \(f_\theta\) minimizing the expected risk on an inaccessible real medical distribution \(\mathcal{D}_{real}\). Formula-Driven Supervised Learning (FDSL) approximates this via a synthetic distribution \(\mathcal{D}_{syn}\), generating labeled volume pairs \((\mathbf{x}, \mathbf{y})\) by sampling \(P(\mathbf{x}, \mathbf{y}) = P(\mathbf{x}|\mathbf{y})P(\mathbf{y})\). Here, the label \(\mathbf{y}\) comprises shapes \(\mathcal{S}\) and spatial transformations \(\mathcal{T}\), represented strictly as sets of voxel coordinates. Previous approaches assume shape structures are statistically independent, factorizing the prior as \(P(\mathbf{y}) = \prod P(S_i)P(T_i)\) with uniform \(T_i\). This unconstrained stochasticity induces a distributional mismatch between \(\mathcal{D}_{syn}\) and \(\mathcal{D}_{real}\).

To bridge this gap, we propose an \textbf{Anatomy-Informed Synthesis} framework redefining \(P(\mathbf{y})\) as a \textit{Constrained Spatial Point Process} governed by a Gibbs distribution conditioned on an anatomical relation graph \(\mathcal{G}=(\mathcal{V}, \mathcal{E})\):
\[
P(\mathbf{y} | \mathcal{G}) = \frac{1}{Z} \exp \left( - \sum_{i \in \mathcal{V}} \psi_{unary}(S_i, T_i) - \sum_{(i,j) \in \mathcal{E}} \psi_{binary}(T_i, T_j) \right)
\]
The prior is anchored in valid human physiology while retaining sufficient variance. Two core energy components achieve this:

\noindent
\textbf{Unary Potential \(\psi_{unary}(S_i, T_i)\):} Encodes a joint shape-position prior. The shape term replaces generic primitives with a non-parametric prior from a lightweight Shape Bank \(\mathcal{B}\) (Sec.~\ref{sec:primitives}). The positional term replaces uniform placement with an anchor-based penalty \(\psi_{unary}^{pos}(T_i) \propto \|\mathbf{t}_{i} - \mathbf{a}_i\|_2\), where the anchor \(\mathbf{a}_i\) is sampled from population statistics \(\mathcal{N}(\mu_i, \Sigma_i)\).
\noindent
\textbf{Binary Potential \(\psi_{binary}(T_i, T_j)\):} Encodes inter-organ topological dependencies. It penalizes physically impossible overlaps and rewards valid boundary contacts, transforming random placement into a structure-aware composition.

Since exact inference under \(P(\mathbf{y}|\mathcal{G})\) is intractable due to the unnormalized \(Z\), we approximate the joint distribution via a sequential greedy strategy:
$
P(\mathbf{y}|\mathcal{G}) \approx \prod_{t=1}^{|\mathcal{V}|} P(T_t | Y_{t-1}, \mathcal{G})
$
where \(Y_{t-1}\) represents the voxel set of previously placed organs. This explicitly bridges our theoretical prior with the tractable candidate ranking generation process detailed in Sec.~\ref{sec:ranking}. Finally, the image \(\mathbf{x} \sim P(\mathbf{x}|\mathbf{y})\) is rendered as contour shells rather than solid volumes, forcing the encoder to capture structural boundaries invariant to texture, while the supervision labels remain dense, filled volumetric masks.

\subsection{Anatomy-Informed Shape Generation}
\label{sec:primitives}

Standard FDSL methods rely on simplistic geometric primitives, which fail to capture the complex boundary curvature of human anatomy. To inject realistic morphological priors, we construct a \textit{Shape Bank} $\mathcal{B}$ derived from a minimal set of real subjects ($K{=}5$) from TotalSegmentator~\cite{Wasserthal2023totalseg}, comprising 32 anatomical classes spanning diverse categories.

To ensure the shapes are clean and reusable, we perform connected component analysis to isolate individual organ instances, followed by bounding-box cropping with a minimal margin. Crucially, we remap the original discrete class indices to a contiguous range $\{1, \dots, C\}$, ensuring a compact label space for the segmentation head.
During synthesis, we apply aggressive geometric augmentations, including random flips, $90^\circ$ rotations, and isotropic scaling within $[0.85, 1.25]$—to expand the effective diversity of $\mathcal{B}$. This ensures the model learns robust, class-agnostic boundary features rather than memorizing specific patient templates.

\subsection{Structure-Aware Sequential Placement}
\label{sec:ranking}

To efficiently solve the placement problem without expensive gradient optimization, we employ a Monte Carlo sampling approach with a candidate ranking mechanism. The placement of an organ $c$ is guided by an \textit{Anatomical Anchor} distribution $\mathcal{A}_c \sim \mathcal{N}(\boldsymbol{\mu}_c, \boldsymbol{\Sigma}_c)$, where $\boldsymbol{\mu}_c$ and $\boldsymbol{\Sigma}_c$ represent the normalized mean centroid and spatial variance derived from the training corpus. 

For each instance, we first sample a specific anchor point $\mathbf{a} \sim \mathcal{A}_c$. We then generate a set of $N$ candidate poses $\Pi = \{\pi_j\}_{j=1}^N$ by sampling random perturbations around $\mathbf{a}$. The optimal pose $\pi^*$ is selected by maximizing a discrete scoring function $S(\pi)$ that aggregates spatial, physical, and topological fidelity:
\begin{equation}
    \pi^* = \operatorname*{arg\,max}_{\pi_j \in \Pi} \left[ S_{spatial}(\pi_j, \mathbf{a}) + S_{phys}(\pi_j, Y_{t-1}) + S_{topo}(\pi_j, Y_{t-1}, \mathcal{G}) \right].
\end{equation}
The \textbf{Spatial Fidelity} term $S_{spatial}$ enforces the generated organ to remain close to the sampled anatomical anchor. It is defined as the negative Euclidean distance $S_{spatial} = - \lambda_{anc} \|\mathbf{t}_{\pi} - \mathbf{a}\|_2$, where $\mathbf{t}_{\pi}$ is the centroid of the candidate mask. This acts as a stochastic tether, ensuring global anatomical correctness while allowing for natural variation.

The \textbf{Physical Constraint} term $S_{phys}$ penalizes unnatural overlaps with the currently occupied volume $Y_{t-1}$. We define this as $S_{phys} = - \lambda_{ovl} \cdot \text{IoU}(\mathbf{m}^\pi, Y_{t-1})$, where $\text{IoU}$ denotes the Intersection-over-Union ratio. Furthermore, we enforce a hard constraint for biologically incompatible pairs (e.g., bone vs. viscera) defined in the exclusion set $\mathcal{E}_{avoid}$. If $\text{IoU}(\mathbf{m}^\pi, Y_{prev}^{(k)}) > \tau_{hard}$ for any $k \in \mathcal{E}_{avoid}$, the candidate is immediately rejected by setting $S(\pi) = -\infty$.

To model inter-organ interactions, we introduce a \textbf{Topological Score} term $S_{topo}$ governed by a relation graph $\mathcal{G}$. This term encourages specific geometric relationships via indicator functions. For containment relationships (e.g., trachea inside lung), we reward candidates where the inclusion ratio exceeds a threshold $\tau_{in}$. Similarly, for adjacency relationships (e.g., liver contacting aorta), we reward candidates that exhibit sufficient boundary contact voxels:
\begin{equation}
    S_{topo} = \sum_{(c, k) \in \mathcal{E}_{in}} \lambda_{in} \mathbb{I}\left( \frac{|\mathbf{m}^\pi \cap Y^{(k)}|}{|\mathbf{m}^\pi|} > \tau_{in} \right) + \sum_{(c, k) \in \mathcal{E}_{adj}} \lambda_{adj} \mathbb{I}\left( |\partial \mathbf{m}^\pi \cap Y^{(k)}| > \nu_{contact} \right).
\end{equation}
By maximizing this composite score, the generator selects the candidate that best balances random variation with anatomical plausibility. The final volume is constructed by sequentially overlaying the selected masks in descending order of their volumes, simulating the occlusion effects inherent in medical imaging.

\section{Experiments}
\label{sec:experiments}
\subsection{Experimental Setup}

\noindent
\textbf{Tasks and datasets:}
We evaluate on both CT and MRI modalities to assess cross-modal generalization. \textit{For CT}, we use BTCV~\cite{landman2015miccai} (multi-organ) and MSD~\cite{antonelli2022medical} Task06 (Lung) and Task09 (Spleen). \textit{For MRI}, we evaluate on MSD Task02 (Heart). This MRI evaluation is particularly important as it tests whether anatomical priors learned from synthetic CT pre-training transfer to a different imaging modality without explicit MRI pre-training data. All datasets are split 80/20 for training/validation, and experiments are conducted offline.

\noindent
\textbf{Data Synthesis Setting:} 
We construct shape bank $\mathcal{B}$ from $K{=}5$ TotalSegmentator~\cite{Wasserthal2023totalseg} subjects with geometric augmentation (flip $p{=}0.5$, $90^\circ$ rotation, scaling $[0.85, 1.25]$). Structure-aware placement uses $N_c{=}40$ candidates per organ with thresholds $r{=}0.7$, $\sigma{=}0.12$, $\tau{=}0.35$, contact minimum $v_{\min}{=}20$, and containment ratio $0.30$. Compatibility weights are $w_o{=}-1.0$, $w_a{=}0.8$, $w_c{=}w_t{=}1.0$.

\noindent
\textbf{Training details:}
During pre-training, we use a ROI size of $96 \times96 \times 96$, a batch size of 4, a learning rate of 0.0002, and a weight decay of 0.00001. The model is optimized using AdamW with a warmup cosine learning rate scheduler. The pre-training dataset contains $5K$ samples, and the number of training iterations is set to $25000$. 
For fine-tuning on the BTCV and MSD datasets, we follow the standard hyperparameter settings used in prior works~\cite{tadokoro2023pre} with a batch size of 4.  All downstream experiments are evaluated using the Dice similarity coefficient.


\begin{table}[tbp]
\centering
\caption{Comparison of segmentation performance on the BTCV dataset. The best results are highlighted in bold.}
\label{tab:btcv_results}

\resizebox{\textwidth}{!}{%
\begin{tabular}{l|c|c|ccccccccccccc}
\toprule
\textbf{Pre-training} & \textbf{PT Num} & \textbf{Avg.} & Spl & RKid & LKid & Gall & Eso & Liv & Sto & Aor & IVC & Veins & Pan & rad & lad \\
\midrule


\multicolumn{16}{l}{\textit{UNETR}} \\

Scratch & 0  & 75.86 
& 93.02 & 89.79 & 90.60 & 51.34 & 71.12 
& 95.02 & 72.90 & 82.95 & 76.71 
& 69.05 & 76.29 & 62.02 & 55.33 \\

PrimGeoSeg~\cite{tadokoro2023pre} & 5K  & 78.90 
& 93.19 & 93.47 & 93.35 & 54.49 & 68.91 
& 95.88 & 78.83 & 87.82 & 82.47 
& 72.73 & 80.08 & 65.06 & 59.35 \\

\rowcolor{lightgray}
Ours & 5K  & \textbf{80.64}
& \textbf{94.23} & \textbf{93.61} & 93.32 & \textbf{62.66} & \textbf{71.76} 
& 95.79 & \textbf{80.51} & \textbf{90.13} & \textbf{84.11} 
& \textbf{73.67} & 79.98 & \textbf{66.49} & \textbf{62.08} \\

\midrule

\multicolumn{16}{l}{\textit{SwinUNETR}} \\
Scratch & 0  & 80.12 & 95.32 & 94.01 & 93.51 & 61.17 & 73.83 & 94.40 & 71.19 & 88.13 & 85.20 & 73.12 & 80.60 & 67.15 & 63.88 \\
PrimGeoSeg~\cite{tadokoro2023pre} & 5K  & 80.08 & 95.25 & 93.76 & 93.79 & 61.23 & 74.47 & 96.04 & 80.64 & 89.39 & 83.91 & 74.27 & 77.83 & 64.13 & 56.36 \\

\rowcolor{lightgray}
Ours & 5K  & \textbf{81.53} 
& 94.39 & \textbf{94.26} & \textbf{93.87} & 59.18 & \textbf{74.92} 
& \textbf{96.67} & 80.00 & \textbf{90.46} & \textbf{85.22} 
& \textbf{74.59} & \textbf{82.88} & \textbf{68.06} & \textbf{65.45}  \\

\bottomrule
\end{tabular}%
}
\end{table}

\begin{table}[t]
\centering
\caption{Comparison of segmentation performance on the MSD dataset across different pre-training strategies. Results are reported for both UNETR and SwinUNETR architectures.}
\label{tab:msd_results}
\resizebox{\textwidth}{!}{
\begin{tabular}{l cc cc cc}
\toprule
\multirow{3}{*}{\textbf{Method}} 
& \multicolumn{5}{c}{\textbf{MSD Dataset}} \\
\cmidrule{2-7}
& \multicolumn{2}{c}{\textbf{Task 02}} 
& \multicolumn{2}{c}{\textbf{Task 06}} 
& \multicolumn{2}{c}{\textbf{Task 09}} \\
\cmidrule{2-3} \cmidrule{4-5} \cmidrule{6-7}
& \textbf{UNETR} & \textbf{SwinUNETR} 
& \textbf{UNETR} & \textbf{SwinUNETR} 
& \textbf{UNETR} & \textbf{SwinUNETR} \\
\midrule
Scratch       & 95.14 & 95.34 & 70.68 & 76.14 & 92.52 & 95.92 \\
PrimGeoSeg   & 95.61 & 95.77  & 78.81 & 80.39 & \textbf{96.30}  & 96.96 \\
\rowcolor{lightgray}
\textbf{Ours}   & \textbf{96.02} & \textbf{95.93}  & \textbf{80.47} & \textbf{81.45}  & 96.20 & \textbf{97.38} \\
\bottomrule
\end{tabular}
}
\end{table}

\begin{table}[tp]
\centering
\caption{Comparison with Self-Supervised Learning (SSL) methods pre-trained on real medical datasets. All models use the SwinUNETR backbone and are fine-tuned on BTCV. (Note: values differ slightly from Table~\ref{tab:btcv_results} due to foreground cropping during fine-tuning, but relative improvements remain consistent.)}
\label{tab:ssl_comparison}
\begin{tabular}{l|c|c|c}
\toprule
\textbf{Method} & \textbf{Pre-train Data} & \textbf{Data Amount} & \textbf{Avg. Dice} \\
\midrule
Scratch & None & 0 & 80.12 \\
\midrule
SwinMM~\cite{wang2023swinmm} & Real (CT) & 5,000 & 76.72 \\
SwinUNETR~\cite{tang2022selfsupervised} & Real (CT) & 5,000 & 80.56 \\
\midrule
PrimGeoSeg & Synthetic & 5,000 & 80.08 \\
\rowcolor{gray!20} \textbf{Ours} & \textbf{Synthetic} & \textbf{5,000} & \textbf{81.51} \\
\bottomrule
\end{tabular}
\end{table}

\subsection{Results and Analysis}
\noindent
\textbf{BTCV Multi-Organ Segmentation.} As shown in Tables~\ref{tab:btcv_results}, for BTCV~\cite{landman2015miccai} dataset, on UNETR, our method achieves $80.64\%$ average Dice, surpassing the scratch baseline by $4.78\%$ and PrimGeoSeg by $1.74\%$. Substantial improvements in structures with weak boundaries (e.g., gallbladder $+11.32\%$, stomach $+7.61\%$) demonstrate that our anatomical priors effectively compensate for ambiguous appearance cues. For SwinUNETR, our method achieves the best overall performance ($81.53\%$), yielding superior accuracy in complex vascular and glandular organs. Notably, PrimGeoSeg fails to surpass the scratch baseline here ($80.08\%$ vs. $80.12\%$), indicating that generic primitives lack the structural complexity needed to initialize sophisticated hierarchical transformers.

\noindent
\textbf{MSD Benchmark Generalization.} 
Our method also exhibits robust cross-dataset and cross-modal transferability, validated on the MSD~\cite{antonelli2022medical} dataset. On Task06 (Lung), our method consistently outperforms both the scratch baseline and the state-of-the-art FDSL method (PrimGeoSeg). Specifically, for UNETR, we achieve a substantial improvement of 9.79\% over training from scratch and 1.66\% over PrimGeoSeg. Similarly, SwinUNETR sees gains of 5.31\% and 1.06\% respectively. Crucially, our method demonstrates effective generalization to MRI (Task02 Heart) despite the synthetic pre-training data being derived exclusively from CT. By achieving 96.02\% (UNETR) and 95.93\% (SwinUNETR), our approach sets a new state-of-the-art, validating that the spatial relationships and topological constraints learned through our structure-aware synthesis are inherently modality-invariant.

\begin{figure*}[t]
    \centering
    \includegraphics[width=0.95\linewidth, height=0.35\linewidth]{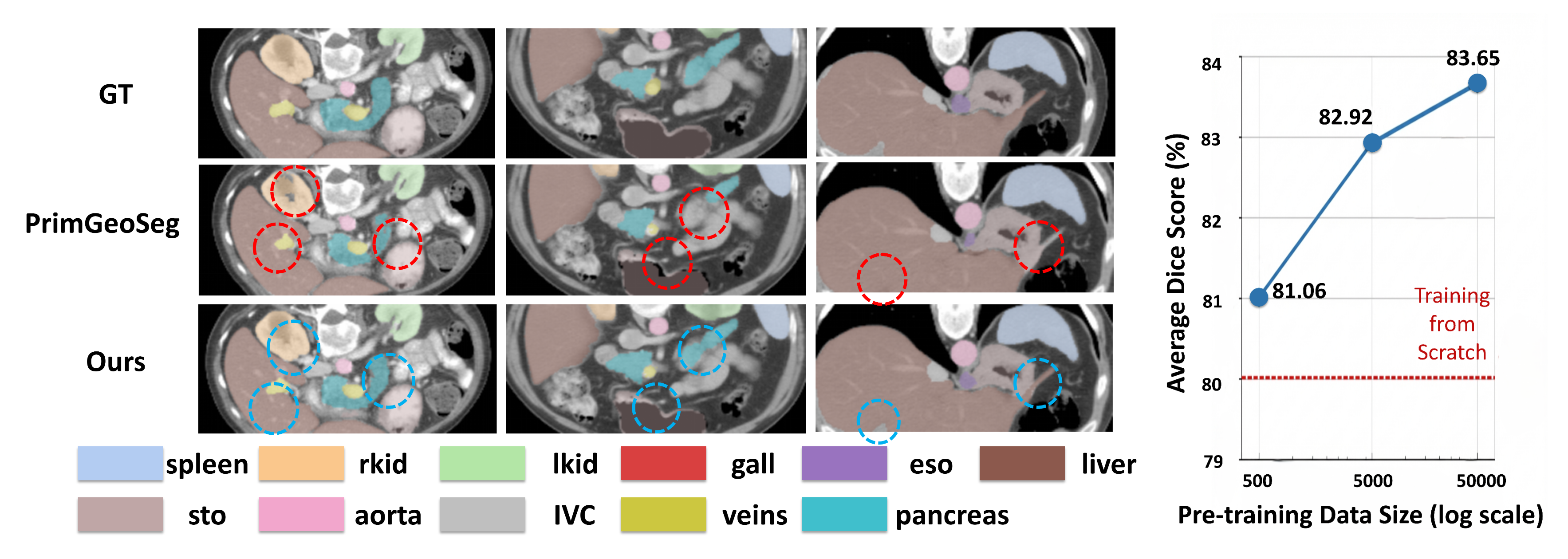}

    \caption{\textbf{Left:} Qualitative comparison results. Red dashes indicate misidentified areas and blue dashes indicate more accurately identified areas. \textbf{Right:} Scaling effect when controlling the generated data number from 500 to 50000.}
    \label{fig:analysis}
\end{figure*}

\noindent
\textbf{Scaling Effect.} We evaluate the impact of synthetic data volume (\(500\), \(5,000\), and \(50,000\)) on UNETR's downstream BTCV performance (Fig.~\ref{fig:analysis}). Our framework consistently outperforms the scratch baseline (\(80.02\%\)), achieving \(81.06\%\), \(82.92\%\), and \(83.65\%\) average Dice, respectively. The marginal gain diminishes as data scales (+\(1.86\%\) from \(500\) to \(5,000\) vs. +\(0.73\%\) from \(5,000\) to \(50,000\)), indicating that \(5,000\) samples offer a practical trade-off between computational cost and accuracy. 

\noindent
\textbf{Comparison with SSL Methods.} We compare our approach against state-of-the-art self-supervised learning methods pre-trained on real medical data (Table~\ref{tab:ssl_comparison}). Using SwinUNETR on BTCV, our anatomy-informed synthetic pre-training (\(81.51\%\)) surpasses both the scratch baseline and PrimGeoSeg. Remarkably, it also exceeds SwinUNETR~\cite{tang2022selfsupervised} pre-trained on \(5,000\) real CT volumes (\(80.56\%\)). In contrast, reconstruction-based SSL like SwinMM~\cite{wang2023swinmm} underperforms the scratch baseline (\(76.72\%\)), suggesting that masked reconstruction objectives may struggle to capture precise semantic features under subtle domain shifts. These results demonstrate that dense, pixel-level anatomical supervision from structured synthetic data is highly effective, offering a privacy-compliant alternative that outperforms unsupervised representation learning on sensitive real scans.

    
%
\section{Conclusion}
In this paper, we presented an Anatomy-Informed Synthetic Supervised Pre-training framework that bridges the infinite scalability of Formula-Driven Supervised Learning (FDSL) with the biological realism of medical scans. By enforcing physiological plausibility and topological constraints through a structure-aware composition strategy, our method significantly outperforms state-of-the-art self-supervised and generic synthetic baselines across multiple CT and MRI benchmarks. Crucially, our work validates a highly promising technical route: \textit{dense, pixel-level synthetic supervision injected with explicit anatomical logic can effectively substitute the reliance on massive, privacy-sensitive real-world datasets.} This paradigm demonstrates that structural priors are more critical than texture reconstruction for medical pre-training, establishing a scalable, data-efficient, and strictly privacy-compliant foundation for training robust 3D medical Transformers in data-limited scenarios.

\bibliographystyle{splncs04}
\bibliography{cite}

@inproceedings{tang2022selfsupervised,
  title     = {Self-Supervised Pre-Training of Swin Transformers for 3D Medical Image Analysis},
  author    = {Tang, Yucheng and Yang, Dong and Li, Wenqi and Roth, Holger R. and Landman, Bennett and Xu, Daguang and Nath, Vishwesh and Hatamizadeh, Ali},
  booktitle = {Proceedings of the IEEE/CVF Conference on Computer Vision and Pattern Recognition},
  year      = {2022},
  pages     = {20730--20740}
}

@inproceedings{hatamizadeh2022unetr,
  title     = {UNETR: Transformers for 3D Medical Image Segmentation},
  author    = {Hatamizadeh, Ali and Tang, Yucheng and Nath, Vishwesh and Yang, Dong and Myronenko, Andriy and Landman, Bennett and Roth, Holger R. and Xu, Daguang},
  booktitle = {Proceedings of the IEEE/CVF Winter Conference on Applications of Computer Vision},
  year      = {2022},
  pages     = {574--584}
}

@article{dosovitskiy2020image,
  title={An image is worth 16x16 words: Transformers for image recognition at scale},
  author={Dosovitskiy, Alexey},
  journal={arXiv preprint arXiv:2010.11929},
  year={2020}
}

@inproceedings{shinoda2023segrcdb,
  title={Segrcdb: Semantic segmentation via formula-driven supervised learning},
  author={Shinoda, Risa and Hayamizu, Ryo and Nakashima, Kodai and Inoue, Nakamasa and Yokota, Rio and Kataoka, Hirokatsu},
  booktitle={Proceedings of the IEEE/CVF international conference on computer vision},
  pages={20054--20063},
  year={2023}
}

@inproceedings{wang2023swinmm,
  title={Swinmm: masked multi-view with swin transformers for 3d medical image segmentation},
  author={Wang, Yiqing and Li, Zihan and Mei, Jieru and Wei, Zihao and Liu, Li and Wang, Chen and Sang, Shengtian and Yuille, Alan L and Xie, Cihang and Zhou, Yuyin},
  booktitle={International conference on medical image computing and computer-assisted intervention},
  pages={486--496},
  year={2023},
  organization={Springer}
}

@inproceedings{landman2015miccai,
  title={Miccai multi-atlas labeling beyond the cranial vault--workshop and challenge},
  author={Landman, Bennett and Xu, Zhoubing and Igelsias, Juan and Styner, Martin and Langerak, Thomas and Klein, Arno},
  booktitle={Proc. MICCAI multi-atlas labeling beyond cranial vault—workshop challenge},
  volume={5},
  pages={12},
  year={2015},
  organization={Munich, Germany}
}

@article{antonelli2022medical,
  title={The medical segmentation decathlon},
  author={Antonelli, Michela and Reinke, Annika and Bakas, Spyridon and Farahani, Keyvan and Kopp-Schneider, Annette and Landman, Bennett A and Litjens, Geert and Menze, Bjoern and Ronneberger, Olaf and Summers, Ronald M and others},
  journal={Nature communications},
  volume={13},
  number={1},
  pages={4128},
  year={2022},
  publisher={Nature Publishing Group UK London}
}

@inproceedings{kataoka2021formula,
  title={Formula-driven supervised learning with recursive tiling patterns},
  author={Kataoka, Hirokatsu and Matsumoto, Asato and Yamada, Ryosuke and Satoh, Yutaka and Yamagata, Eisuke and Inoue, Nakamasa},
  booktitle={Proceedings of the IEEE/CVF International Conference on Computer Vision},
  pages={4098--4105},
  year={2021}
}

@inproceedings{kataoka2022replacing,
  title={Replacing labeled real-image datasets with auto-generated contours},
  author={Kataoka, Hirokatsu and Hayamizu, Ryo and Yamada, Ryosuke and Nakashima, Kodai and Takashima, Sora and Zhang, Xinyu and Martinez-Noriega, Edgar Josafat and Inoue, Nakamasa and Yokota, Rio},
  booktitle={Proceedings of the IEEE/CVF conference on computer vision and pattern recognition},
  pages={21232--21241},
  year={2022}
}

@inproceedings{takashima2023visual,
  title={Visual atoms: Pre-training vision transformers with sinusoidal waves},
  author={Takashima, Sora and Hayamizu, Ryo and Inoue, Nakamasa and Kataoka, Hirokatsu and Yokota, Rio},
  booktitle={Proceedings of the IEEE/CVF Conference on Computer Vision and Pattern Recognition},
  pages={18579--18588},
  year={2023}
}

@inproceedings{tadokoro2023pre,
  title={Pre-training auto-generated volumetric shapes for 3d medical image segmentation},
  author={Tadokoro, Ryu and Yamada, Ryosuke and Kataoka, Hirokatsu},
  booktitle={Proceedings of the IEEE/CVF Conference on Computer Vision and Pattern Recognition},
  pages={4740--4745},
  year={2023}
}

@inproceedings{nakamura2023pre,
  title={Pre-training vision transformers with very limited synthesized images},
  author={Nakamura, Ryo and Kataoka, Hirokatsu and Takashima, Sora and Noriega, Edgar Josafat Martinez and Yokota, Rio and Inoue, Nakamasa},
  booktitle={Proceedings of the IEEE/CVF International Conference on Computer Vision},
  pages={20360--20369},
  year={2023}
}

@misc{he2021mae,
      title={Masked Autoencoders Are Scalable Vision Learners}, 
      author={Kaiming He and Xinlei Chen and Saining Xie and Yanghao Li and Piotr Dollár and Ross Girshick},
      year={2021},
      eprint={2111.06377},
      archivePrefix={arXiv},
      primaryClass={cs.CV},
      url={https://arxiv.org/abs/2111.06377}, 
}

@misc{xie2022simmim,
      title={SimMIM: A Simple Framework for Masked Image Modeling}, 
      author={Zhenda Xie and Zheng Zhang and Yue Cao and Yutong Lin and Jianmin Bao and Zhuliang Yao and Qi Dai and Han Hu},
      year={2022},
      eprint={2111.09886},
      archivePrefix={arXiv},
      primaryClass={cs.CV},
      url={https://arxiv.org/abs/2111.09886}, 
}

@article{Wasserthal2023totalseg,
   title={TotalSegmentator: Robust Segmentation of 104 Anatomic Structures in CT Images},
   volume={5},
   ISSN={2638-6100},
   url={http://dx.doi.org/10.1148/ryai.230024},
   DOI={10.1148/ryai.230024},
   number={5},
   journal={Radiology: Artificial Intelligence},
   publisher={Radiological Society of North America (RSNA)},
   author={Wasserthal, Jakob and Breit, Hanns-Christian and Meyer, Manfred T. and Pradella, Maurice and Hinck, Daniel and Sauter, Alexander W. and Heye, Tobias and Boll, Daniel T. and Cyriac, Joshy and Yang, Shan and Bach, Michael and Segeroth, Martin},
   year={2023},
   month=sep }

@InProceedings{Tang2022SwinUNETR,
    author    = {Tang, Yucheng and Yang, Dong and Li, Wenqi and Roth, Holger R. and Landman, Bennett and Xu, Daguang and Nath, Vishwesh and Hatamizadeh, Ali},
    title     = {Self-Supervised Pre-Training of Swin Transformers for 3D Medical Image Analysis},
    booktitle = {Proceedings of the IEEE/CVF Conference on Computer Vision and Pattern Recognition (CVPR)},
    month     = {June},
    year      = {2022},
    pages     = {20730-20740}
}

@misc{hatamizadeh2022swinunetr,
      title={Swin UNETR: Swin Transformers for Semantic Segmentation of Brain Tumors in MRI Images}, 
      author={Ali Hatamizadeh and Vishwesh Nath and Yucheng Tang and Dong Yang and Holger Roth and Daguang Xu},
      year={2022},
      eprint={2201.01266},
      archivePrefix={arXiv},
      primaryClass={eess.IV},
      url={https://arxiv.org/abs/2201.01266}, 
}

@misc{zhou2019models,
      title={Models Genesis: Generic Autodidactic Models for 3D Medical Image Analysis}, 
      author={Zongwei Zhou and Vatsal Sodha and Md Mahfuzur Rahman Siddiquee and Ruibin Feng and Nima Tajbakhsh and Michael B. Gotway and Jianming Liang},
      year={2019},
      eprint={1908.06912},
      archivePrefix={arXiv},
      primaryClass={eess.IV},
      url={https://arxiv.org/abs/1908.06912}, 
}

@misc{wu2024voco,
      title={VoCo: A Simple-yet-Effective Volume Contrastive Learning Framework for 3D Medical Image Analysis}, 
      author={Linshan Wu and Jiaxin Zhuang and Hao Chen},
      year={2024},
      eprint={2402.17300},
      archivePrefix={arXiv},
      primaryClass={eess.IV},
      url={https://arxiv.org/abs/2402.17300}, 
}

\end{document}